\pdfoutput=1

\documentclass[11pt]{article}

\usepackage[final]{latex/acl}
\usepackage{hyperref}
\usepackage{times}
\usepackage{latexsym}

\usepackage[T1]{fontenc}

\usepackage[utf8]{inputenc}

\usepackage{microtype}
\usepackage{multirow}
\usepackage{inconsolata}

\usepackage{graphicx}

\usepackage{url}
\usepackage{amsmath,amssymb,amsfonts}
\usepackage{algorithmic}
\usepackage{booktabs}
\usepackage{makecell}
\usepackage{multirow}
\usepackage{tabularx}
\usepackage{array}
\usepackage{textcomp}
\usepackage{xcolor}
\usepackage{comment}
\usepackage{mdframed}
\usepackage{pifont} 

\definecolor{lightgray}{gray}{0.9}  
\title{Enhancing Speech-to-Speech Dialogue Modeling with End-to-End Retrieval-Augmented Generation}

\author{
Pengchao Feng\textsuperscript{1},
Ziyang Ma\textsuperscript{1}, 
Wenxi Chen\textsuperscript{1}, 
Yao Li\textsuperscript{3},
Sheng Wang\textsuperscript{3}, 
Kai Yu\textsuperscript{1}, 
Xie Chen\textsuperscript{1,2} \thanks{\ Corresponding Author.}
\\  \textsuperscript{1}MoE Key Lab of Artificial Intelligence, Jiangsu Key Lab of Language Computing,
\\ X-LANCE Lab, School of Computer Science, Shanghai Jiao Tong University, Shanghai, China \\
    \textsuperscript{2}Shanghai Innovation Institute 
    \textsuperscript{3}Shanghai Aviation Electric Co., Ltd 
}

\begin{document}

\maketitle

\begin{abstract}
End-to-end speech-to-speech (S2S) dialogue systems have recently garnered increasing research attention for their lower latency and more natural integration of nonverbal cues such as emotion and speaker identity. However, these systems face key challenges, particularly in incorporating external knowledge, a capability commonly addressed by Retrieval-Augmented Generation (RAG) in text-based large language models (LLMs). The core difficulty lies in the modality gap between input speech and retrieved textual knowledge, which hinders effective integration of information. To address this issue, we propose a novel end-to-end RAG framework that directly retrieves relevant textual knowledge from speech queries. Experimental results demonstrate that our method significantly improves the performance of end-to-end S2S dialogue systems while achieving higher retrieval efficiency. Although the overall performance still lags behind the SOTA cascaded models, our framework offers a promising direction for enhancing knowledge integration in end-to-end S2S systems. Our code and dataset are released. \footnote{The code is available in \url{https://github.com/the-bird-F/GLM-Voice-RAG}. The speech dataset is available in \url{https://huggingface.co/datasets/the-bird-F/HotpotQA_RGBzh_speech}.}
\end{abstract}

\section{Introduction}
The release of GPT-4o~\cite{gpt4o}, a unified model capable of comprehending and generating text, speech, and visual modalities within a single neural architecture, represents a significant milestone in the development of Multimodal Large Language Models (MLLMs). MLLMs are defined as large language model-based architectures that can perceive~\cite{dosovitskiy2020image, QwenAudio}, reason over~\cite{zhang2023multimodal, ma2025audio}, and generate~\cite{AnyGPT, du2023lauragpt} multimodal information \cite{MLLM}, with representative models including AnyGPT \cite{AnyGPT} and Qwen-Omni \cite{xu2025qwen2}. Among various MLLM paradigms, end-to-end speech-to-speech (S2S) dialogue models, such as Mini-Omni~\cite{xie2024mini}, LLaMA-Omni~\cite{fang2024llama}, and SLAM-Omni~\cite{chen2024slam}, constitute an important subclass, aiming to directly respond to speech input with speech output through a single integrated framework. However, due to the nature of the modality gap between input speech and retrieved textual knowledge, these models lack the capability to retrieve external knowledge documents and face limitations in timely knowledge updates and domain-specific information acquisition, hindering their performance in question-answering tasks.

Retrieval-Augmented Generation (RAG) \cite{rag}, which retrieves relevant information from an external source and incorporates it as contextual input to enhance response accuracy, effectively addresses the limitation of text-based LLMs in accessing external knowledge \cite{RAGsurvey}. However, in end-to-end speech-to-speech dialogue models, directly retrieving textual documents is often ineffective due to the difficulty of aligning input speech with the corresponding textual knowledge. A straightforward solution is to leverage Automatic Speech Recognition (ASR) to convert speech into text and then use the recognized text for retrieval. While this approach can be effective, high-accuracy ASR models, such as Whisper \cite{whisper}, introduce additional latency, which poses a critical limitation for real-time vertical spoken assistants.

Therefore, balancing retrieval accuracy and latency remains a critical challenge in building high-performance voice-based retrieval assistants. Given that RAG relies on encoding textual information into a vector space for indexing and retrieval, we propose an end-to-end approach that embeds speech queries and textual content into a shared representation space, enabling direct speech-to-text retrieval without needing ASR. This approach significantly reduces retrieval latency and streamlines the pipeline. 

In our experiments with speech-to-speech dialogue models, the proposed end-to-end RAG approach reduces retrieval latency to one-fourth of that incurred by the ASR-based cascaded RAG models. Specifically, on spoken question-answering tasks, our model achieves an effective improvement in answer accuracy over the baseline without retrieval augmentation (eg, $20\%$ improvement on HotpotQA \cite{hotpotQA}), although performing lower than the cascade models employing ASR-based cascade retrieval. These findings highlight the effectiveness of our approach and establish a promising design paradigm for future research on multimodal RAG.

In summary, our key contributions are as follows:
\begin{itemize}
    \item We propose an effective end-to-end RAG framework for speech-to-speech dialogue models, which is able to retrieve textual knowledge with input speech directly.
    \item We conduct a comprehensive evaluation on multilingual benchmarks, which not only demonstrates that the end-to-end retrieval method achieves a 4× speedup in retrieval, but also underscores the critical role of semantic alignment in retrieval accuracy.
    \item To promote reproducibility and further research, we open-source the code and dataset used in our experiment. 
\end{itemize}

\begin{figure*}[htbp]
    \centering
    \includegraphics[width=1\textwidth]{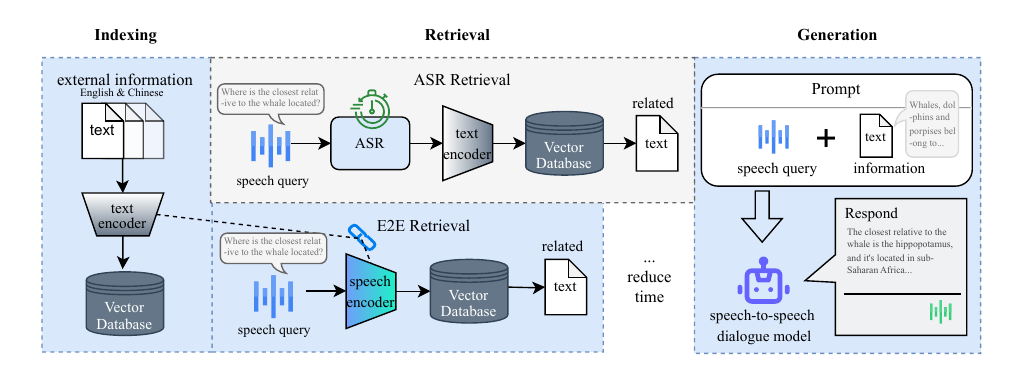} 
    \caption{Our proposed framework of end-to-end RAG for speech-to-speech dialogue model. 1) Indexing stage, where text information is divided into chunks and encoded index. 2) Retrieval stage, where we compare ASR-based cascade retrieval, which transcribes speech to text before performing text retrieval, and end-to-end retrieval, which encodes speech queries into the same vector space as texts for direct retrieval. 3) Generation stage, where the end-to-end S2S dialogue model generates responses.}
    \label{fig:model}
\end{figure*}

\section{Related Work}

\subsection*{RAG for Speech-to-Speech Dialogue Models}
ReSLM \cite{wang2024retrieval} introduces the concept of a speech retriever for the speech-language model SLM \cite{wang2023slm}, which leverages retrieved audio context for more grounded responses and lays the foundation for this area.

ReCap \cite{ghosh2024recap} and SRAG \cite{yang2024srag} extend speech retrieval techniques to audio captioning and spoken language understanding, respectively.

SpeechRAG \cite{min2024speech} and SEAL \cite{SEAL} are among the first works to explore the use of a shared embedding space for enabling speech-to-text retrieval. And next WavRAG \cite{wavrag} leverages the audio encoding capabilities of Qwen-Audio to generate semantically rich speech embeddings for retrieval. 

Our work is contemporaneous with approaches such as SpeechRAG, SEAL, and WavRAG, all driven by the recent surge in spoken dialogue systems and aimed at universal cross-modal RAG. What distinguishes our work is that it not only highlights the trade-offs between ASR-based RAG systems and an end-to-end retrieval system, but also presents multilingual experiments demonstrating superior generalization across languages.

\subsection*{Speech-Text Alignment}
Speech-text alignment is a foundational task in speech processing, underpinning key applications such as automatic speech recognition (ASR) and text-to-speech (TTS) synthesis.

Recent models such as SeamlessM4T \cite{seamlessm4t} and Whisper \cite{whisper} have achieved accurate and robust performance across diverse languages and acoustic conditions. However, in speech-to-text retrieval tasks, high ASR accuracy does not necessarily lead to better performance, as ASR introduces latency, and full transcriptions are often unnecessary.

SONAR \cite{sonar} is a multi-encoder embedding framework that constructs a sentence embedding space from text and uses it as a teacher model to train a speech encoder aligned with the corresponding textual representation. The alignment enables direct retrieval of textual information from speech queries. 

\section{Method}

\subsection{System overview}
Figure \ref{fig:model} illustrates our proposed end-to-end Retrieval-Augmented Generation system architecture. The system comprises three integrated stages: indexing, retrieval, and generation. Unlike conventional text-based RAG systems, our framework uniquely supports both speech and text modalities across all processing stages.

\subsection{Indexing}
This stage aims to construct a database from existing text documents.

\begin{equation}
    C = \{ (\varphi_{\text{text}}(d_i),\ d_i) | \  D \to  \{d_i\} \}
    \label{i2}
\end{equation}

As formula \ref{i2}, the process involves three key steps: (1) the text documents $D$ are divided into chunks $\{d_i\}$. (2) Each chunk is embedded using the text encoder $\varphi_{\text{text}}$ from a multimodal embedding model, which maps different modalities into a shared vector space.  In this work, we adopt SONAR \cite{sonar} as the encoder function $\varphi$, which comprises modality-specific encoders for text ($\varphi_{\text{text}}$) and speech ($\varphi_{\text{speech}}$), respectively. 
(3) Finally, the encoded indexes and their corresponding text chunks are stored in the database $C$. 

\subsection{Retrieval}
The retrieval stage aims to extract relevant textual information in response to the spoken query.

\begin{equation}
    d_{\mathrm{r}} = \operatorname*{arg\,top\text{-}k}_{d\in C} \, \mathrm{sim}(\varphi_{\text{speech}}(q),\varphi_{\text{text}}(d))
    \label{r1}
\end{equation}

As formula~\ref{r1}, we employ the speech encoder $\varphi_{\text{speech}}$, derived from the same embedding model as $\varphi_{\text{text}}$, to encode the entire spoken query, thereby obviating the need for speech tokenization. Then the retrieval module computes the similarity between the query embedding and the indexed data chunks, and returns the top-$k$ most similar chunks, which constitute the relevant retrieved information.

By eliminating the need to convert speech to text, the retrieval process reduces the time and enhances overall efficiency.

\subsection{Generation}
In the final stage, the retrieved information is combined with the original spoken query and fed into the end-to-end speech-to-speech dialogue model. This enables the model to generate responses grounded in factual knowledge, improving its relevance and practicality for real-world applications.

To integrate the user query with the retrieved information and guide the model toward generating factually grounded responses, we use the prompt provided by the langchain \cite{langchain}, which is provided in the appendix.

\begin{table*}[ht]
\small
    \centering
    \begin{tabular}{ c c c c c c c c}
        \toprule
        \multirow{2}{*}{Method} & \multirow{2}{*}{Embedding} &  \multicolumn{3}{c}{\textbf{HotpotQA (en)}} &  \multicolumn{3}{c}{\textbf{RGB (zh)}} \\ 
        \cmidrule(lr){3-5} \cmidrule(lr){6-8}
        & & retrieval.t $\downarrow$ & retrieval.f1 $\uparrow$ & answer.acc $\uparrow$ & retrieval.t$\downarrow$ & retrieval.f1 $\uparrow$ & answer.acc $\uparrow$\\
         \midrule
         w/o RAG  & -  & - & - & 0.27 & - & - & 0.17\\
         \midrule
        \multirow{3}{*}{ASR RAG}  
        & OpenAI & ~1.24 s$^*$ & 0.27 & 0.48 & ~1.23 s$^*$ & 0.42 & 0.68 \\
        & BCE & 0.41 s & 0.25 & 0.48 &  0.34 s& 0.44 & 0.69 \\
        & M-E5  & 0.43 s & \textbf{0.28} & \textbf{0.52} &  0.33 s & \textbf{0.47} &  \textbf{0.72}\\
        \midrule
        E2E RAG & SONAR  & \textbf{0.08 s} & 0.24 & 0.43  & \textbf{0.07 s} & 0.31  & 0.54\\
        \midrule
         \multirow{3}{*}{\textcolor{gray}{Orcale RAG}}
        & \textcolor{gray}{OpenAI} &  \textcolor{gray}{-} & \textcolor{gray}{0.28} & \textcolor{gray}{0.50} & \textcolor{gray}{-} & \textcolor{gray}{0.42} & \textcolor{gray}{0.69}  \\
         & \textcolor{gray}{BCE}  & \textcolor{gray}{-} & \textcolor{gray}{0.25} & \textcolor{gray}{0.48}  & \textcolor{gray}{-} & \textcolor{gray}{0.41} & \textcolor{gray}{0.68}\\
         & \textcolor{gray}{M-E5}   & \textcolor{gray}{-}  & \textcolor{gray}{0.28} & \textcolor{gray}{0.53} &  \textcolor{gray}{-} & \textcolor{gray}{0.46} & \textcolor{gray}{0.73}\\
         \midrule
         \textcolor{gray}{Facts}  & \textcolor{gray}{-}  & \textcolor{gray}{-} & \textcolor{gray}{-} & \textcolor{gray}{0.69} & \textcolor{gray}{-} & \textcolor{gray}{-} & \textcolor{gray}{0.94}\\
        \bottomrule
    \end{tabular}
    \caption{Performance on two benchmarks: HotpotQA (English) and RGB-zh (Chinese). 
     "w/o RAG" and "Facts" represent retrieval-free and oracle-knowledge settings, respectively. We evaluate ASR RAG and Oracle RAG methods with different embedding models (OpenAI, BCE, M-E5).  
     $^*$: may be affected by the network environment. } 
    \label{tab: results}
\end{table*}

\section{Experiments}

\subsection{Dataset}

\textbf{HotpotQA} \cite{hotpotQA} is a diverse and interpretable question-answering dataset that necessitates multi-hop reasoning. The test-distractor dataset, comprising 7,405 questions and answers, was selected for evaluation. We call Azure's text-to-speech API \footnote{\url{https://learn.microsoft.com/en-us/azure/ai-services/speech-service}} to synthesize the corresponding speech data.

\textbf{RGB} \cite{RGB} is an RAG benchmark, available in both English and Chinese, which constructs QA data based on the latest news from 2024. For the experiment, we selected the Chinese test set and used the Azure API to generate the corresponding speech data.

\subsection{Baseline systems}
We construct four system groups for comparison, with their configs provided in the appendix. 

\begin{itemize}
    \item \textbf{w/o RAG} This baseline system allows the dialogue model to answer questions directly, without retrieving external knowledge, thereby reflecting the model's intrinsic knowledge.
        
    \item \textbf{Facts} This topline system bypasses retrieval and directly accesses the relevant information. It aims to evaluate the comprehension, analysis, and reasoning capabilities of the dialogue model itself.

    \item \textbf{ASR RAG} These systems convert the input speech query into text using an ASR system, and then apply text-based RAG. They represent the most straightforward approach to speech-text retrieval and are designed to contrast the differences between cascading RAG and end-to-end RAG in terms of retrieval efficiency and answer accuracy. 
    
    \item \textbf{Oracle RAG} These systems typically assume that ASR transcriptions are error-free, allowing RAG to be applied directly on the transcribed text without correction. They serve as accuracy benchmarks for text-based RAG to identify issues in implementation or modal alignment.
    
\end{itemize}

\subsection{Result}
The results are shown in Table \ref{tab: results}. We examine the effect of the model from two aspects: retrieval efficiency and answer accuracy of the whole dialogue system.

\textbf{Retrieval Efficiency} For each retrieval, we measured the time taken to retrieve information and the similarity of the retrieved information to the ground-truth facts (i.e., the information that can effectively answer the question), which was measured using the F1 score between two text strings. The results show that our end-to-end retrieval system can reduce the retrieval time of the cascade system (ASR RAG) to 1/4 on both the English and Chinese datasets (reducing 0.4s to 0.08s). And the retrieval accuracy of our model is  1-4 percentage points lower than that of the cascade model.  

\textbf{Answer Accuracy} For the whole speech-to-speech dialogue system, we use the evaluation metric (Recall, covered Exact Match) provided by the benchmarks to measure the accuracy of the system's responses. From the results, we can see that our system has a significant improvement over the native dialogue model without RAG ( $\sim 20\%$ improvement on HotpotQA and $\sim 43\%$  improvement on RGB-zh); However, it is lower than the model using text directly and the model using high-accuracy ASR retrieval ($9\%$ lower than the system using the best embedding on HotpotQA and $\sim 15\%$ lower than the system using RGB-zh). 

The result shows that in the scenario where response rate is not emphasized, the ASR-based cascade model still needs to be used, which has higher answer accuracy. However, considering some scenarios that require rapid response, such as E-commerce promotion and concert ticket buying, end-to-end RAG is a good trade-off between retrieval efficiency and answer accuracy, which has higher practical value.

\subsection{Ablation study}
The additional ablation experiment, where the speech encoder was removed and replaced with an ASR retrieval pipeline using Whisper and the SONAR text encoder to perform retrieval, or an alternative speech-text alignment tool CLAP \cite{elizalde2023clap}, shows the following results:

\begin{table}[ht]
\centering
\small
\begin{tabular}{ccccc}
\toprule
\multirow{2}{*}{Speech Encoder} & \multicolumn{2}{c}{\textbf{HotpotQA}} & \multicolumn{2}{c}{\textbf{RGB}} \\
\cmidrule(lr){2-3} \cmidrule(lr){4-5}
& re.f1$\uparrow$& an.acc$\uparrow$ & re.f1$\uparrow$& an.acc$\uparrow$\\
\midrule
SONAR(ours)  & 0.24 & 0.43 & 0.31 & 0.54 \\
\ding{55} & 0.25 & 0.44 & 0.30 & 0.56 \\
CLAP  & 0.14 & 0.39 & 0.16& 0.42 \\
\bottomrule
\end{tabular}
\caption{Ablation study evaluating the contribution of SONAR speech encoder by comparing its presence or absence, which replaces it with an ASR pipeline (Whisper and SONAR text encoder, marked \ding{55}), or CLAP.}
\label{ablation}
\end{table}

As shown in Table \ref{ablation},  when the SONAR text encoder is used for text embeddings, using or removing the SONAR speech encoder yields nearly identical accuracy, which indicates that the accuracy gap between ASR RAG and E2E RAG is not due to the inclusion of the speech encoder, but rather due to the limitations of the SONAR text encoder. 

From another perspective, using the SONAR speech encoder with a text decoder for speech recognition tasks \cite{sonar} shows that SONAR performs comparably to Whisper v3, suggesting that the SONAR speech encoder introduces minimal additional error.

Additionally, we analyzed failed retrieval cases (Appendix A.4), which show that E2E retrieval using SONAR retrieves fewer semantically relevant segments, particularly when no lexical overlap exists. This supports that the shared embedding space from SONAR lacks strong semantic alignment compared to more advanced encoders like BCE. 

Table \ref{ablation} also demonstrates that replacing SONAR with CLAP leads to a drop in accuracy. The reason is that CLAP is optimized for speech–text alignment but less effective in preserving fine-grained text semantic alignment, as it does not construct a highly semantically coherent embedding space. This result again confirms that \textbf{semantic alignment of the shared vector space is a key factor for improving retrieval accuracy.}

And in SONAR training, the text encoder serves as the teacher, and the speech encoder is optimized via MSE loss to align its output with the corresponding text embedding.
Therefore, using more powerful text encoders (e.g., BCE, M-E5, and OpenAI embedding) as the teacher model could theoretically lead to better-trained speech encoders with a stronger semantic vector space, and also enable higher accuracy in E2E RAG systems without sacrificing efficiency.

\section{Conclusion}
In this work, we highlight the efficiency bottleneck in cross-modal retrieval, spark discussion around the trade-off between latency and accuracy, and contribute an open-sourced, multilingual end-to-end retrieval augmented generation framework to facilitate further research. The proposed method leverages a pretrained embedding model with multiple encoders, allowing both speech and text to be indexed within the same vector space, which enables direct retrieval of relevant textual information from speech queries. The experimental results demonstrate that our method significantly enhances retrieval speed, which is an efficient and scalable framework for end-to-end speech-to-speech dialogue systems. Furthermore, we point out that the key factor in improving retrieval accuracy is enhancing the semantic alignment of the shared vector space, which provides guidance for future research.

\section*{Limitations}
Despite the promising results of our proposed end-to-end RAG framework for spoken dialogue systems, there are still notable limitations that warrant further investigation. 

At the speech-text alignment level, one major limitation is the retrieval accuracy constrained by the suboptimal performance of the SONAR embedding model used in our system. This limitation is especially critical when high-precision semantic understanding is required for information retrieval. 

At the RAG level, another significant limitation is that our current approach utilizes only the native RAG model for experimentation. It remains unclear whether adopting more advanced RAG frameworks, such as graph RAG, could further improve the performance of end-to-end speech-to-speech dialogue systems. Future work could investigate the potential benefits of integrating these frameworks to better understand their impact on system performance.

\section*{Acknowledgements}
This work was supported by the National Natural Science Foundation of China  (No. U23B2018 and No. 62206171), Shanghai Municipal Science and Technology Major Project under Grant 2021SHZDZX0102, and Yangtze River Delta Science and Technology Innovation Community Joint Research Project (2024CSJGG01100).

\bibliography{latex/custom}

\clearpage
\appendix
\section{Appendix}
\label{sec:appendix}
All datasets and tools are intended for research use, and our use complies with that purpose.
\subsection{Experiment Config}
The experiments are conducted on a single NVIDIA A100-PCIE-40GB GPU.

For TTS, we used 'JennyNeural', an American female voice, to generate English speech data, and 'XiaoxiaoNeural', a Mandarin-speaking female voice, for Chinese speech data.

And we adopt three text embedding models as a function $\varphi$ of our index construction. 1) OpenAI embedding \cite{GPT4} which comes from OpenAI; 2) BCEmbedding base v1 \cite{bce} which comes from YouDao by Wangyi. 3) Multilingual-E5-text-embeddings-large (M-E5) \cite{multi} which comes from Microsoft. 

Furthermore, we employed the end-to-end speech-to-speech dialogue model GLM-4-Voice \cite{GLM-4-Voice} as the generator for our case study, and used the top-1 response as the answer. GLM-4-Voice \footnote{\url{https://github.com/THUDM/GLM-4-Voice}}, developed by Zhipu AI, is an end-to-end speech-to-speech model capable of directly understanding and generating speech in both Chinese and English. It supports real-time voice interaction and allows for the modification of voice attributes based on user commands, making it highly suitable for our experimental needs. 

The top 4 relevant chunks were retrieved in all of our experiments.

\subsection{Prompt Frame}
The main part of the prompt is provided by the langchain \cite{langchain} retrieval tool, which is proven to be reliable. The assistant prompt is used to prompt the large model to transcribe in streaming mode.
\begin{mdframed}[backgroundcolor=lightgray, linewidth=0pt, innertopmargin=5pt, innerbottommargin=5pt]
\textbf{Prompt:} \\
Human: You are an assistant for question-answering tasks. Use the following pieces of retrieved context to answer the question. If you don't know the answer, just say that you don't know. Use three sentences maximum and keep the answer concise.\\
Question: \{speech query\}\\
Context: \{retrieved text\}\\
assistant: streaming\_transcription
\end{mdframed}

The system prompt adopts the GLM-4-Voice prompt format during instruction fine-tuning to ensure the model can better leverage the responses learned through fine-tuning.
\begin{mdframed}[backgroundcolor=lightgray, linewidth=0pt, innertopmargin=5pt, innerbottommargin=5pt]
\textbf{Prompt:} \\
 system: The User will provide you with a speech instruction. Do it step by step. First, think about the instruction and respond in an interleaved manner, with 13 text tokens followed by 26 audio tokens.
\end{mdframed}

\subsection{ASR}
We adopt Whisper-large-v3 \cite{whisper} as the ASR model.

The model, when transcribing the speech generated by the Azure Speech Service, introduces a certain level of error, with a word error rate (WER) of 13.0\% on English data and a character error rate (CER) of 9.5\% on Chinese data. For reference, this version of the Whisper model reports a WER of 9.3\% on English and a CER of 12.8\% on Mandarin Chinese based on the Common Voice 15 dataset \cite{common} when evaluated independently \cite{whisper}.

\subsection{Experiment Cases}
\subsubsection*{Success cases}
\begin{mdframed}[backgroundcolor=lightgray, linewidth=0pt, innertopmargin=5pt, innerbottommargin=5pt]
\textbf{Question:} \\
 What science fantasy young adult series, told in first person, has a set of companion books narrating the stories of enslaved worlds and alien species?\\ \textbf{Ground Truth:} \\
 Animorphs \\
\textbf{Useful Information:} \\
'The Hork-Bajir Chronicles is the second companion book to the "Animorphs" series, written by K. A. Applegate.', ' With respect to continuity within the series, it takes place before book 23, "The Pretender", although the events told in the story occur between the time of "The Ellimist Chronicles" and "The Andalite Chronicles".', ' The book is introduced by Tobias, who flies to the valley of the free Hork-Bajir, where Jara Hamee tells him the story of how the Yeerks enslaved the Hork-Bajir, and how Aldrea, an Andalite, and her companion, Dak Hamee, a Hork-Bajir, tried to save their world from the invasion.', 'Animorphs is a science fantasy series of young adult books written by Katherine Applegate and her husband Michael Grant, writing together under the name K. A. Applegate, and published by Scholastic.', ' It is told in first person, with all six main characters taking turns narrating the books through their own perspectives.'\\
\textbf{Prediction:} \\
 It sounds like you're describing 'The Animorphs' series. It's a science fantasy young adult book series where the main characters can transform into animals. The series explores themes like friendship, family, and fighting for justice. If you're looking for more suggestions, I'd be happy to help! \\
\textbf{Result: True} \\
\end{mdframed}

\begin{mdframed}[backgroundcolor=lightgray, linewidth=0pt, innertopmargin=5pt, innerbottommargin=5pt]
\textbf{Question:}  \\
Are the Laleli Mosque and Esma Sultan Mansion located in the same neighborhood?
\\ \textbf{Ground Truth:} \\
 No \\
\textbf{Useful Information:} \\
'The Laleli Mosque (Turkish: "Laleli Camii, or Tulip Mosque" ) is an 18th-century Ottoman imperial mosque located in Laleli, Fatih, Istanbul, Turkey.', 'The Esma Sultan Mansion (Turkish: "Esma Sultan Yalısı" ), a historical yalı (English: waterside mansion ) located at Bosphorus in Ortaköy neighborhood of Istanbul, Turkey and named after its original owner Esma Sultan, is used today as a cultural center after being redeveloped.'\\
\textbf{Prediction:} \\ Laleli Mosque and Esma Sultan Mansion are not in the same neighborhood. Laleli Mosque is in the Fatih district, while Esma Sultan Mansion is in the Ortaköy neighborhood, which is part of the Beyoğlu district. They're both in Istanbul, but they're quite a distance apart.
  \\
\textbf{Result: True} \\
\end{mdframed}

\subsubsection*{Failure cases}
\begin{mdframed}[backgroundcolor=lightgray, linewidth=0pt, innertopmargin=5pt, innerbottommargin=5pt]
\textbf{Question :} \\ The director of the romantic comedy "Big Stone Gap" is based in what New York city?
\\ \textbf{Ground Truth:} \\ Greenwich Village
 \\
\textbf{Useful Information:} \\ 'Big Stone Gap is a 2014 American drama romantic comedy film written and directed by Adriana Trigiani and produced by Donna Gigliotti for Altar Identity Studios, a subsidiary of Media Society.', 'Adriana Trigiani is an Italian American best-selling author of sixteen books, television writer, film director, and entrepreneur based in Greenwich Village, New York City.'
\\
\textbf{Prediction:} \\  The director of "Big Stone Gap" is likely not based in New York City, as the film is set in a small town in the United States. However, if you are referring to a specific character or context within the film, please provide more details for me to better assist you.
  \\
\textbf{Result: False} \\
\end{mdframed}

\begin{mdframed}[backgroundcolor=lightgray, linewidth=0pt, innertopmargin=5pt, innerbottommargin=5pt]
\textbf{Question:} \\
What government position was held by the woman who portrayed Corliss Archer in the film Kiss and Tell?\\ \textbf{Ground Truth:} \\
Chief of Protocol \\
\textbf{Useful Information:} \\
'Kiss and Tell is a 1945 American comedy film starring then 17-year-old Shirley Temple as Corliss Archer.', "Shirley Temple Black (April 23, 1928 – February 10, 2014) was an American actress, singer, dancer, businesswoman, and diplomat who was Hollywood's number one box-office draw as a child actress from 1935 to 1938.", ' As an adult, she was named United States ambassador to Ghana and to Czechoslovakia and also served as Chief of Protocol of the United States.'\\
\textbf{Prediction:} \\
In "Kiss and Tell," Corliss Archer was portrayed by Shirley Temple. However, I don't know which government position she held in real life or if she held any at all. Temple was best known for her film career, and she never held a government position. \\
\textbf{Result: False} \\
\end{mdframed}
\end{document}